\documentclass{article}

\usepackage{microtype}
\usepackage{graphicx}    
\usepackage{caption}     
\usepackage{subcaption}

\usepackage{booktabs} 

\usepackage{amsmath,amsfonts,bm}

\def\secref#1{section~\ref{#1}}

\def\eqref#1{equation~\ref{#1}}

\def\1{\bm{1}}

\DeclareMathAlphabet{\mathsfit}{\encodingdefault}{\sfdefault}{m}{sl}
\SetMathAlphabet{\mathsfit}{bold}{\encodingdefault}{\sfdefault}{bx}{n}

\usepackage{enumitem}

\usepackage[utf8]{inputenc} 

\usepackage[T1]{fontenc}    

\usepackage[colorlinks,citecolor=forestgreen]{hyperref}       
\usepackage{url}            
\usepackage{nicefrac}       

\usepackage{booktabs}
\usepackage{array}
\usepackage{multirow}
\usepackage{xspace}
\usepackage[table]{xcolor}
\usepackage{pgfplots}

\usepackage{listings}
\usepackage[most]{tcolorbox}
\usepackage{adjustbox}
\usepackage{array}

\usepackage{float}

\newcommand{\methodname}{\textit{Dr.SoW}\xspace}

\DeclareMathOperator{\template}{T}

\lstdefinelanguage{Java}{
    keywords={class, int, public, return, while, null},
    keywordstyle=\color{blue}\bfseries,
    ndkeywords={Node},
    ndkeywordstyle=\color{teal}\bfseries,
    identifierstyle=\color{black},
    sensitive=true,
    commentstyle=\color{gray}\itshape,
    stringstyle=\color{orange}\bfseries,
    morecomment=[l]//,
    morestring=[b]"
}

\usepackage[accepted]{icml2025}

\usepackage{amsmath}
\usepackage{amssymb}
\usepackage{mathtools}
\usepackage{amsthm}

\usepackage[capitalize,noabbrev]{cleveref}
\usepackage[textsize=tiny]{todonotes}

\theoremstyle{plain}

\theoremstyle{definition}

\theoremstyle{remark}

\pgfplotsset{compat=1.18}
\renewcommand{\eqref}[1]{(\ref{#1})}

\icmltitlerunning{Dr. SoW: Density Ratio of Strong-over-weak LLMs for Reducing the Cost of Human Annotation in Preference Tuning}

\begin{document}

\twocolumn[
\icmltitle{Dr. SoW: Density Ratio of Strong-over-weak LLMs for Reducing the Cost of Human Annotation in Preference Tuning}

\begin{icmlauthorlist}
\icmlauthor{Guangxuan Xu}{RH}
\icmlauthor{Kai Xu}{RH}
\icmlauthor{Shivchander Sudalairaj}{RH}
\icmlauthor{Hao Wang}{RH}
\icmlauthor{Akash Srivastava}{RH}

\end{icmlauthorlist}

\icmlaffiliation{RH}{RedHat AI Innovation}

\icmlcorrespondingauthor{Guangxuan Xu}{gxxu@redhat.com}

\icmlkeywords{Machine Learning, ICML}

\vskip 0.3in
]
\printAffiliationsAndNotice{}  

\begin{abstract}
Preference tuning relies on high-quality human preference data, which is often expensive and time-consuming to gather. In this paper, we introduce \methodname (Density Ratio of Strong over Weak) a cost-effective method that eliminates the reliance for human annotation by leveraging off-the-shelf LLMs for preference data annotation. 
\methodname uses the log-density ratio between a better-aligned and a less-aligned LLM as a reward signal. 
We evaluate \methodname across 221 different LLM pairs and empirically find a strong correlation between the performance gap of the paired models and the quality of the reward signal. This insight provides a practical guideline for selecting LLMs for data annotation. 
Additionally, we introduce an end-to-end pipeline that customizes reward functions based on user query domains. Without fine-tuning, it improves accuracy on domain-specific evaluations.

With a pair of Mistral-7B models, \methodname achieves a RewardBench score of 82.6, outperforming the best trained reward functions from same model class and demonstrating competitive performance against SoTA models in Safety (91.0) and Reasoning (88.0) domains. 
Further, we preference-tune \textit{Llama-3-8B-Instruct} using data annotated by \methodname. Our approach pushes Llama-3-8B to achieve a 37.4\% ($+$15.1\%) win rate on ArenaHard and a 40.7\% ($+$17.8\%) win rate on length-controlled AlpacaEval 2.0. 
\end{abstract}

\begin{figure}[t]
\begin{center}
\includegraphics[width=0.45\textwidth]{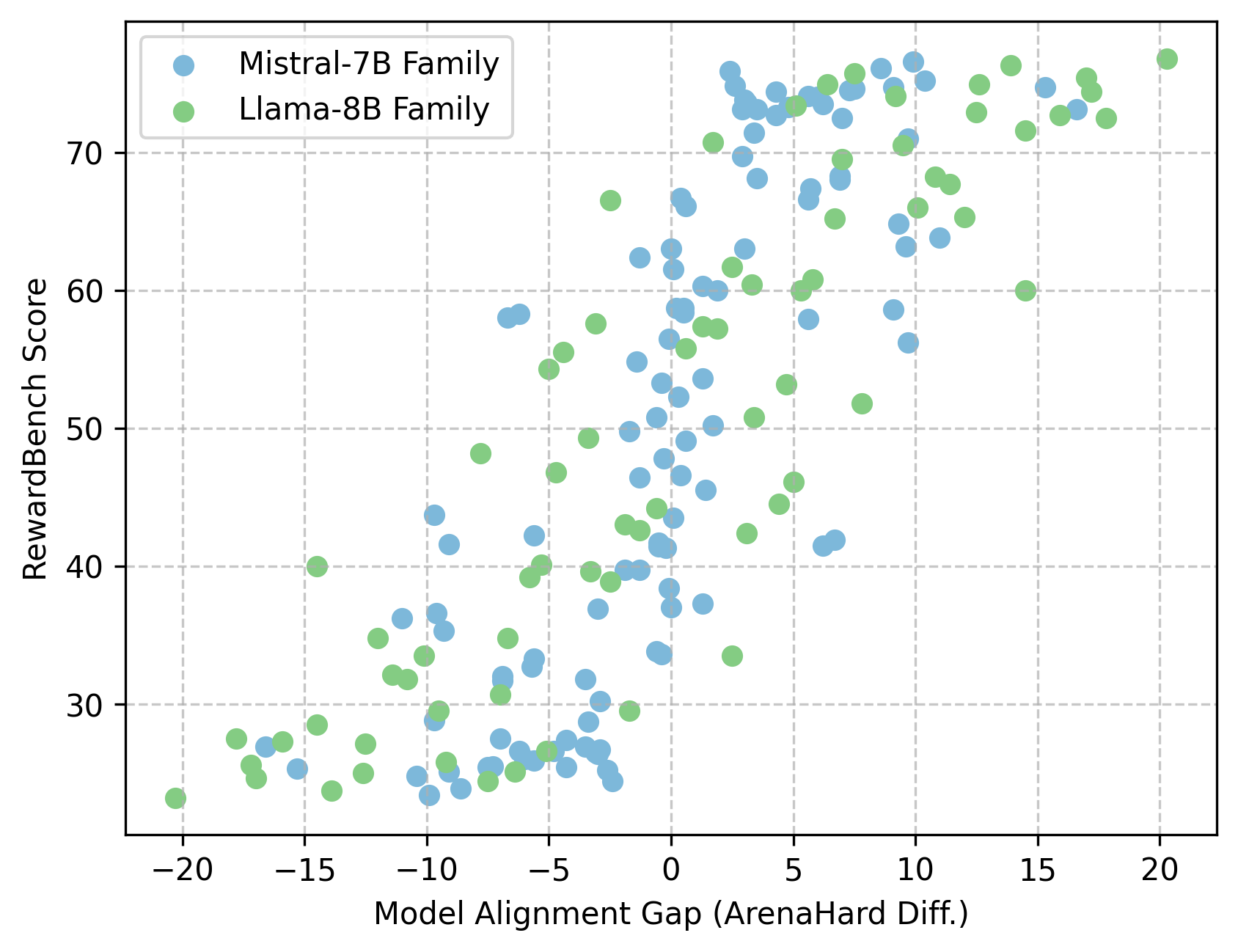}
\end{center}
\caption{
We analyze how different model pairs $(\pi_{\text{strong}}, \pi_{\text{weak}})$ impact the quality of the reward signal provided by \eqref{eq:weak_strong_ratio}. Each point represents one of 221 unique model pairs: 100 Llama-8B pairs (green) and 121 Mistral-7B pairs (blue). The x-axis denotes the alignment gap between $\pi_{\text{strong}}$ and $\pi_{\text{weak}}$, measured by ArenaHard scores, while the y-axis represents reward signal quality, measured by RewardBench scores. We observe a strong correlation between model alignment gap and reward signal quality, indicating that practitioners should pair a well-aligned $\pi_{\text{strong}}$ with a less-aligned $\pi_{\text{weak}}$ when using \eqref{eq:weak_strong_ratio} as a reward signal.}
\label{fig:dot-plot}
\vspace{-1.5em}
\end{figure}

\section{Introduction}
Preference tuning has advanced the capabilities of large language models (LLMs), but this progress relies on high-quality human preference data which is both costly and time-consuming to gather. Cutting-edge models are aligned with curated, quality-controlled human preference data, typically provided by specialized companies. While effective, this approach limits broader adoption due to prohibitive costs and limited transparency in data collection~\citep{Wang2024HelpSteer2OD}. AI-feedback solutions are emerging as an alternative---either through a trained reward model~\citep{Dong2024RLHFWF} or proprietary LLM-as-a-judge~\citep{Cui2023UltraFeedbackBL}. However, training such reward models still rely on costly initial human preference data; and LLM-as-a-judge approaches introduce licensing restrictions that generally prevent commercial use when using proprietary models.

We introduce \methodname (Density Ratio of Strong-over-Weak), an automatic labeling method that not only drastically reduces manual costs in preference annotation, but also is comparable or beats proprietary model-as-a-judge method and trained reward models in reward accuracy and preference alignment outcome. Our method leverages the log-density ratio between a better-aligned and a less-aligned model to annotate preference data, offering a flexible approach applicable to any off-the-shelf open-source LLMs. Through extensive experiments across 221 model combinations (Figure~\ref{fig:dot-plot}), we provide design guidelines for selecting model pairs. Our findings reveal that a larger alignment gap between models enhances the reward signal for preference annotation, a principle we term the ``Strong-over-Weak Hypothesis''. 
Our approach generalizes the DPO implicit reward, which restricts model pair selection to post-DPO and pre-DPO models \citep{Chen2024BootstrappingLM}. We demonstrate that by selecting a model pair with more significant alignment gap, the reward signal defined by \methodname could outperform the DPO implicit reward (Figure~\ref{fig:iter-dpo}). This flexibility allows models trained with diverse objectives---including SFT, RRHF, SLiC-HF, ORPO, SimPO, KTO, and IPO---to be used for data annotation. Moreover, our results offer actionable design guidelines for practitioners seeking to optimize reward function quality.

Customizing the reward function for data annotation is crucial to ensuring alignment with domain-specific needs. For instance, safety annotation may prioritize risk minimization and policy compliance, whereas code annotation might emphasize correctness and readability, and math annotation could focus on logical consistency and precision. A generic and one-size-fits-all reward function fails to capture these nuanced requirements. A common approach involves fine-tuning reward models for each domain, but this process is costly due to the need for domain-specific data collection and model training~\citep{Ji2024PKUSafeRLHFTM, wang2024self}. We streamline this process by introducing an end-to-end pipeline that identifies the domain of each user query and customizes the density-ratio reward function to prioritize relevant preference criteria. Specifically, \methodname employs an adaptive router to classify queries into domains such as chat, reasoning, and safety. It then applies domain-specific instructions and in-context learning examples to refine preference criteria. In this way, we customize a density-ratio reward function from a general preference signal to domain-specific annotators. Experimental results show that adaptively customized density-ratio rewards significantly enhance both overall and domain-specific reward signal quality.

In summary, our main contributions are:
\begin{itemize}[leftmargin=1em]

    \item \textbf{Cost-effective preference annotation.} We introduce a scalable, cost-effective pipeline for preference data annotation. By leveraging the density ratio of off-the-shelf LLMs as a reward function, it drastically reduces the reliance on human annotation and allows for domain customization of reward without requiring additional data or fine-tuning. This automated annotation process can drastically lower the cost of human labeling, while also minimizing the expertise and computational resources traditionally needed for training reward models.

    \item \textbf{Broader model choice and better reward signals.} \methodname enables the use of any open-source or in-house models for preference data annotation. It goes beyond existing methods that rely on proprietary models or special model pairs for data annotation. We formalize the strong-over-weak hypothesis, which provides a principled guideline for selecting LLMs to produce a stronger reward signal. We observe that certain model pairs yield higher-quality reward functions than the DPO implicit reward.

    \item \textbf{Strong alignment performance.} We provide an end-to-end preference data annotation pipeline and validate it through extensive experiments. With a pair of Mistral-7B models, \methodname achieves a RewardBench score of 82.6, outperforming the best trained reward functions from same model class and demonstrating competitive performance against SoTA models in Safety (91.0) and Reasoning (88.0) domains. Further, we preference tune \textit{Llama-3-8B-Instruct} using data annotated by \methodname. Our approach pushes Llama-3-8B to achieve a 37.4\% ($+$15.1\%) win rate on ArenaHard and a 40.7\% ($+$17.8\%) win rate on length-controlled AlpacaEval 2.0. This outperforms model aligned with data from SoTA-level reward classifiers, proving our approach is both cost-effective and highly effective.

\end{itemize}

\section{Background}
Prior studies~\citep{Lin2024OnTL, Chen2024BootstrappingLM} has explored using implicit reward from direct policy optimization~\citep[DPO;][]{rafailov2023directpreference} for preference data annotation. 
DPO is a preference-based fine-tuning method that does not require (explicit) reward modeling. Instead, it directly optimizes a policy language model $\pi_{\theta}$ using a reference model $\pi_{\text{ref}}$, typically an SFT model. The policy $\pi_{\theta}$ is initialized as $\pi_{\text{ref}}$, and the (implicit) reward function being optimized in DPO is:
\begin{align}
\label{eq::dpo_reward}
    r_{\text{DPO}}(x,y) = \beta \log \frac{\pi_{\theta}(y | x)}{\pi_{\text{ref}}(y | x)} + \beta \log(Z(x))
\end{align}
where $x$ is the prompt, $y$ is the answer,  $\beta$ is a temperature hyperparameter and $Z(x)$ is a normalization constant. Ignoring the normalization constant, this reward function is the log-density ratio between a specific model pair: the policy model being optimized and its reference model.

A series of works~\citep{Lambert2024RewardBenchER, Lin2024OnTL, Chen2024BootstrappingLM} explored leveraging the implicit reward function of DPO to annotate preference data. They proposed selecting a post-DPO model and a pre-DPO model to define a reward function. By definition, the pre-DPO model is the reference model (typically a SFT model) used during DPO training. Given a prompt $x$ and two responses, $y_1$ and $y_2$, the response with the higher reward is labeled as preferred, while the other is labeled as dispreferred.

\section{Method}
We study two research questions critical to density-ratio-based reward function design. First, we investigate whether alternative model pairs can produce stronger signals compared to the DPO implicit reward (\secref{sec:method-strong-over-weak}). Our experiments reveal a positive correlation between the alignment gap of model pairs (measured by the ArenaHard score) and the effectiveness of the reward function (evaluated through the RewardBench score). By increasing the gap in human alignment levels, we observe that certain model pairs yield a stronger reward signal  than the DPO implicit reward. Second, we investigate whether we can further refine density-ratio reward based on domain characteristics of annotation data (\secref{sec:method-customization}). We show that conditioning the density ratio with domain-related instructions and exemplars significantly improve overall and in-domain reward signal quality without requiring additional training.

\begin{figure*}[t]
\begin{center}
\begin{subfigure}[t]{0.45\textwidth}
        \centering
        \includegraphics[width=\textwidth]{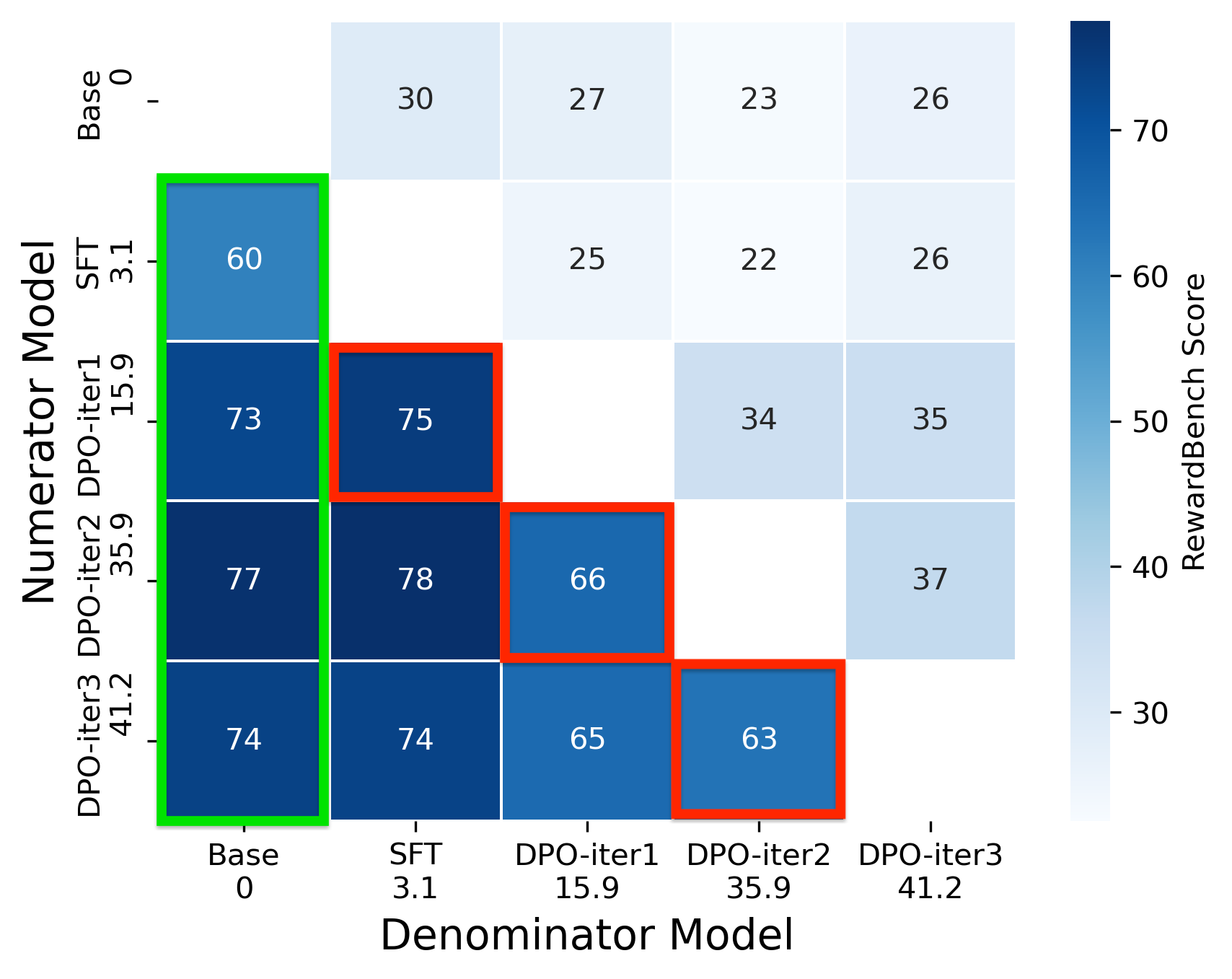}
        \caption{Llama-3-8B}
        \label{fig:iter-dpo-llama}
    \end{subfigure}
    \hfill
    \begin{subfigure}[t]{0.45\textwidth}
        \centering
        \includegraphics[width=\textwidth]{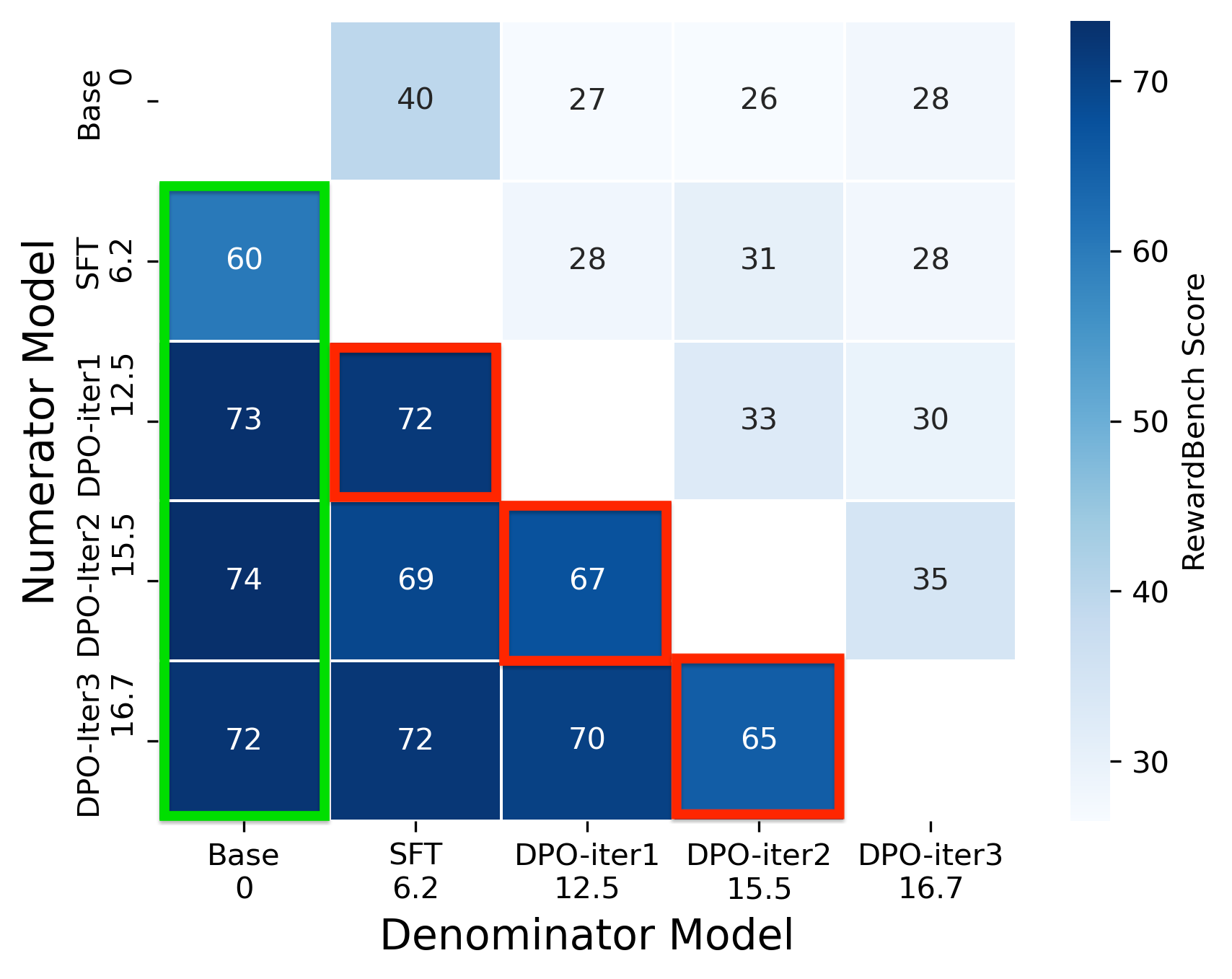}
        \caption{Mistral-7B}
        \label{fig:iter-dpo-mistral}
    \end{subfigure}
\end{center}
\caption{
Density ratio reward from different pairing combinations, with y-axis the numerator model, and x-axis denominator model. The five models chosen in each model family are sorted by their human-aligned level measured by ArenaHard. According to DPO implicit reward theory, models along the diagonal (red-outlined cells) theoretically yield optimal rewards, pairing models before and after DPO training. However, empirical results indicate that using the Base model as the denominator consistently yields higher scores (green-outlined cells), motivating our strong-over-weak density ratio reward function. }
\label{fig:iter-dpo}
\end{figure*}

\subsection{Density-ratio Reward Functions}
\label{sec:method-strong-over-weak}

\paragraph{Motivation} We explore constructing density-ratio-based reward function with various pairings of LLMs. At first glance, one might assume that the DPO model and its reference model would be the optimal pair for this purpose. To examine this hypothesis, we conduct an experiment using online iterative DPO~\citep{Xiong2023IterativePL, Xu2023SomeTA, Swamy2024AMA} trained models from the Mistral and Llama-3 families. 
The key ideas of online iterative DPO training are: (1) the reference model is updated at each iteration (i.e., $\pi_{\text{ref}} = \pi_{\theta_{t-1}}$), and (2) the training data is also updated iteratively by sampling responses from $\pi_{\theta_{t-1}}(\cdot \mid x)$ and annotated with an external reward function.

In this online iterative DPO setting, the policy model $\pi_{\theta_t}$ at iteration $t$ uses the previous iteration's policy model $\pi_{\theta_{t-1}}$ as its reference. According to the implicit DPO reward theory, one might expect the density ratio between $\pi_{\theta_t}$ and $\pi_{\theta_{t-1}}$ to provide an optimal reward function. However, Figure~\ref{fig:iter-dpo} shows that using weaker models---such as the base or SFT models---as the denominator in \eqref{eq:weak_strong_ratio}, instead of $\pi_{\theta_{t-1}}$, produces significantly better reward functions as evaluated by RewardBench. This finding indicates that the DPO implicit reward is empirically suboptimal compared with simply choosing weaker models in the denominator of \eqref{eq:weak_strong_ratio}, implication of which motivates us to propose the ``Strong-over-Weak Hypothesis''.

\paragraph{Reward Function Design}

We use the following reward function to annotate preference data. 
\begin{equation}
r(x, y) = \log \frac{\pi_{\text{strong}}(y \mid x)}{\pi_{\text{weak}}(y \mid x)}. 
\label{eq:weak_strong_ratio}
\end{equation}

Here $\pi_\text{strong}$ and $\pi_\text{weak}$ are two off-the-shelf LLMs from the same model family with $\pi_\text{strong}$ outperforming $\pi_\text{weak}$ across all dimensions of human preference, such as safety, correctness, and relevance.

\textbf{Strong-over-Weak Hypothesis} We conduct extensive experiments using $221$ distinct model pairs to construct various reward functions in \eqref{eq:weak_strong_ratio} and evaluate their quality on RewardBench. Our findings reveal a strong correlation between the alignment gap of $\pi_\text{strong}$ and $\pi_\text{weak}$ and the effectiveness of the reward function, as quantified by the RewardBench score. As shown in Figure~\ref{fig:dot-plot}, achieving an effective reward function in \eqref{eq:weak_strong_ratio} with a high RewardBench score requires a substantial human-alignment difference between $\pi_\text{strong}$ and $\pi_\text{weak}$. We refer to this insight as the ``Strong-over-Weak Hypothesis'', which serves as a guiding principle for constructing density-ratio-based reward function as in \eqref{eq:weak_strong_ratio}. Our experiments span a range of models, including base, SFT, SimPO, KTO, ORPO, going beyond post-DPO and pre-DPO models (see Figure~\ref{fig:mistral-optimization-rb} for details). We summarize our key observations below.

\begin{itemize}[leftmargin=1em]
    \item We recommend using a weak model for the denominator in (\ref{eq:weak_strong_ratio}) that has not been fine-tuned on human preference data, such as an \textit{SFT} or \textit{base} model. For the numerator, a stronger model that aligns more closely with human preferences (e.g., AlpacaEval2.0 or ArenaHard benchmarks) should be used. This approach maximizes the performance gap, often leading to better performance of the reward function.

    \item We recommend using both strong and weak models from the same model family. If the weak model is an SFT model, we suggest using a strong model that has been preference-tuned from this SFT model. This approach ensures that when leveraging existing benchmarks (e.g., AlpacaEval 2.0 or ArenaHard) to evaluate the performance gap in human preference alignment, potential confounding factors, such as differing inductive biases between unrelated models, are minimized.
    
\end{itemize}

\begin{figure*}[h]
    \centering
    \begin{tcolorbox}
        \small
        You are a helpful AI assistant. You follow the following guidelines when answering user questions.

        \vspace{0.5em}
        \textbf{1. Answer Constructive, Clear Questions} \\
        - Provide an answer when the user asks for factual information, constructive advice, or help with personal growth. Focus on offering practical, positive guidance.

        \vspace{0.5em}
        \textbf{2. Recognize Jokes, Puns, and Fictional Contexts} \\
        - Respond playfully when the question references humor, games, movies, or fictional scenarios. Acknowledge the fictional nature while keeping the tone light.

        \vspace{0.5em}
        \textbf{3. Avoid Answering Harmful, Illegal, or Malicious Questions} \\
        - Do not engage if the question promotes harm, illegal activities, or unethical behavior. Politely but firmly refuse to provide an answer, while keeping the response respectful.

        \vspace{0.5em}
        \textbf{4. Handle Sensitive Topics with Empathy} \\
        - Respond with care to questions about mental health, personal relationships, or emotionally charged situations. Acknowledge the user's feelings, and offer general advice or suggest professional resources.
    \end{tcolorbox}
    \caption{Instruction with detailed criterion to define preference in Safety domain. This prompt outlines key principles to ensure constructive, empathetic, and safe responses.}
    \label{fig:safety}
    \vspace{-1.5em}
\end{figure*}

\subsection{Reward Function Customization}
\label{sec:method-customization}

Human preferences are multi-dimensional (e.g., safety, trustworthiness, reliability,  faithfulness)~\citep{bai2022constitutionalai,Wang2024HelpSteer2OD, naseem-etal-2024-grounded}, and an effective reward function should adapt its criteria according to the specific domain requirements. For example, a chatbot explaining corporate vacation policies should emphasize faithfulness to company policy and the accuracy of its responses, rather than focusing on aspects like conversational style or user engagement. However, vanilla log-density ratio reward function provides a single, aggregated reward signal, merging various, potentially conflicting preference aspects.

We introduce \methodname, which offers customized preference criterion for annotating samples from different domains through the use of instructions and in-context-learning (ICL) examples. Each domain has its own sets of instructions and ICL examples, and we ensure diversity by preparing multiple ICL demonstrations, sampling one randomly for each instruction. Formally, for each original user prompt $x$, we inject ICL examples and domain-specific instructions $\template(x)$ to guide the annotation toward relevant preference dimensions. This is equivalent to adapting the reward function into the following form, incorporating $\template(x)$ before applying the log-density ratio for annotation.

\begin{equation}
r_{\text{\methodname}}(x, y) = \log \frac{\pi_{\text{strong}}(y \mid \template(x), x)}{\pi_{\text{weak}}(y \mid \template(x), x)}.
\label{eq:log_ratio-prompted}
\end{equation}
To automate annotation, we introduce a domain router that identifies the most relevant domain for each user query. We then apply appropriate preference criteria to each example in the annotation set. For instance, a sensitive query is routed to a Safety expert, while a math or coding query goes to a Math/Code expert. We use the Mixtral 8x7B Instruct v0.1 model \citep{mixtral} with zero-shot prompting to classify prompts into pre-defined categories (e.g., safety, reasoning, chat) based on a system prompt and task description.

We provide a pool of domain-specific in-context examples and instructions, such as those in Figure~\ref{fig:safety_in_context},~\ref{fig:math_in_context},~\ref{fig:reason_in_context} (Appendix~\ref{app:auto-prompt}). They serve as both demonstrative and descriptive tools to help refine the reward model's preference criterion. Example templates we used can be found in Figure~\ref{fig:safety}. For domains like safety, instructions should include guidelines on how to avoid risky outcomes, while in domains like math, demonstrating the preference criterion through examples may be more effective. These instructions provide high-level guidance by defining overarching principles that shape the reward function's preferences during data annotation. 

If users wish to automatically discover preference criteria for their target domain, we provide an automated pipeline for generating preference instruction prompts. This reduces manual effort in prompt engineering and enhances the accessibility of our approach. Inspired by \citet{DOosterlinck2024InContextLF}, our prompt tuning method iteratively constructs the prompt based on an initial prompt and the user-provided evaluation dataset; see details in Appendix~\ref{app:auto-prompt}. It achieves performance comparable to manually crafted prompts (see Table~\ref{tab:prompt_performance}).

\section{Experiments}
\subsection{Strong-Over-Weak Reward Annotation}
\label{sec:exp-weak-to-strong}

\paragraph{Setup}
We collect model pairs, $\pi_{\text{strong}}$ and $\pi_{\text{weak}}$, from two families—Mistral and Llama. These models exhibit distinct levels of human alignment, as measured by ArenaHard \citep{Li2024FromCD}, a benchmark demonstrated to yield the highest correlation and separability with real human judgments in ChatArena.
We then assess the density ratio reward function of distinct model combinations through RewardBench~\citep{Lambert2024RewardBenchER}. Each sample in RewardBench consists of a human-verified pair: one chosen and one rejected response. The reward function then assigns annotations by comparing the density ratio scores of these two responses. The final score reflects the accuracy of the reward function's predictions against human-annotated ground truth.
Our experiment includes base models, supervised fine-tuning (SFT) models, as well as models optimized through different preference-tuning algorithms.

\begin{figure*}[t]
\begin{center}
\begin{subfigure}[t]{0.48\textwidth}
        \centering
        \includegraphics[width=\textwidth]{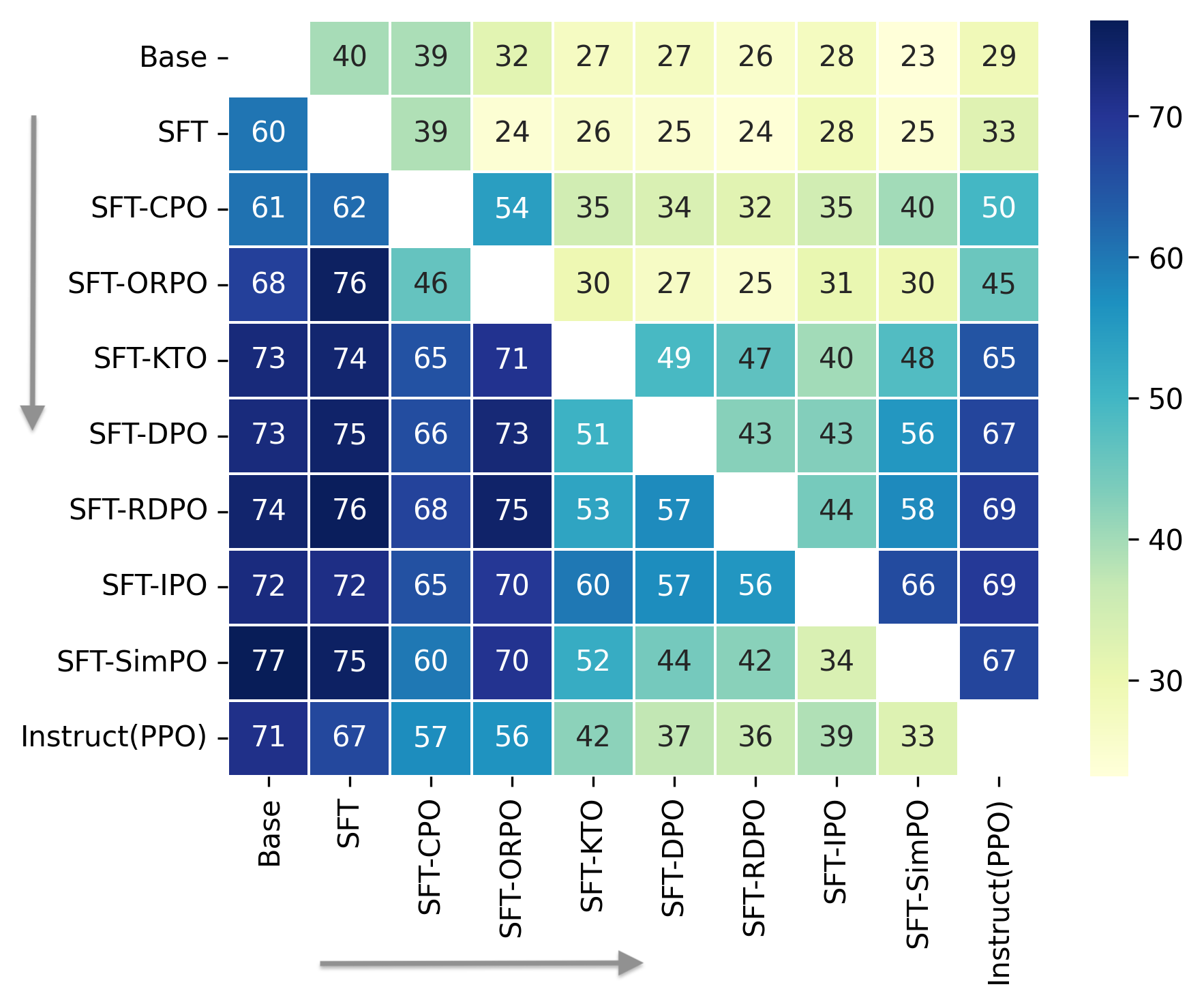}
        \caption{Llama-3-8B Family}
        \label{fig:heatmap-llama}
        \vspace{-1em}
    \end{subfigure}
    \hfill
    \begin{subfigure}[t]{0.48\textwidth}
        \centering
        \includegraphics[width=\textwidth]{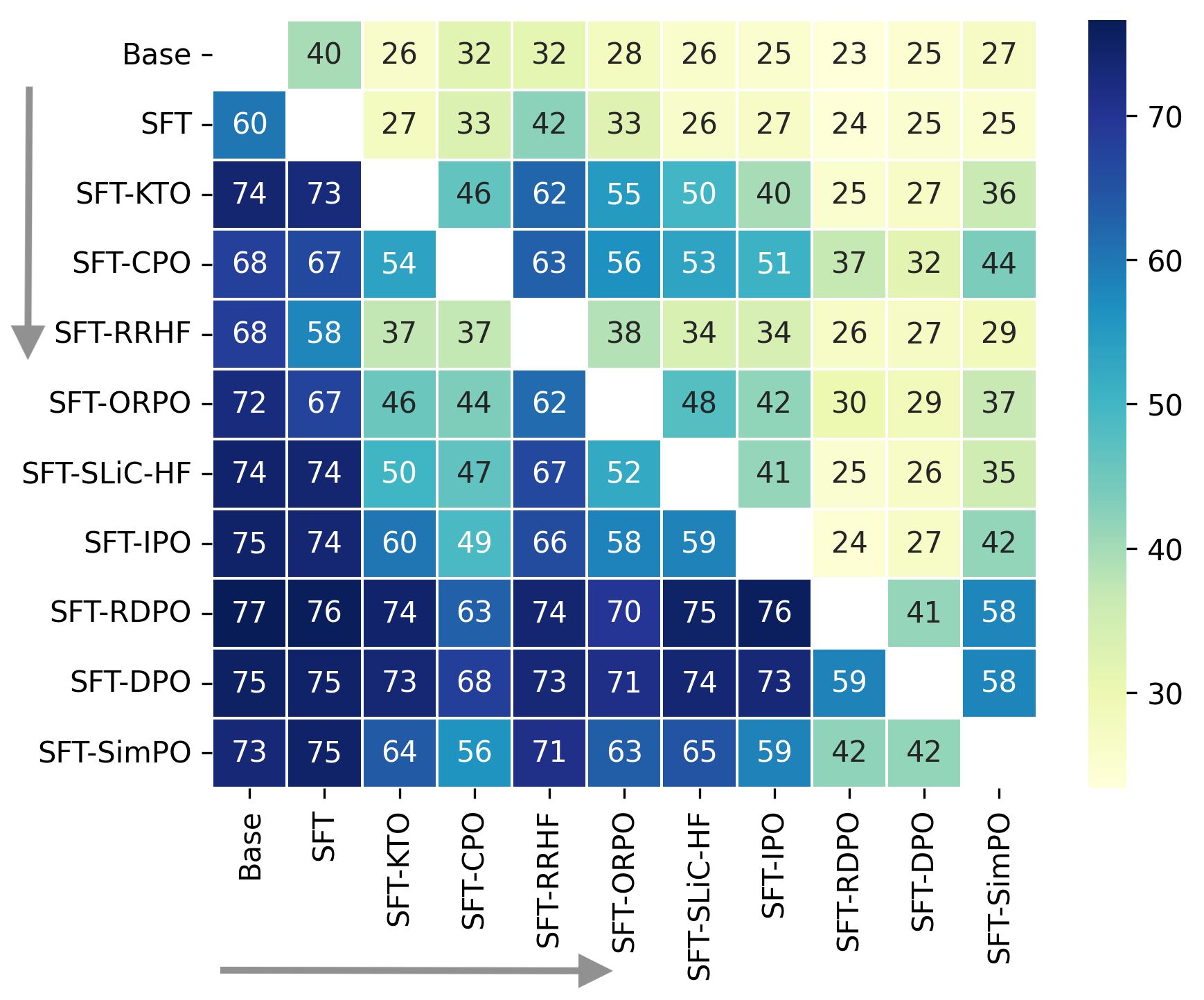}
        \caption{Mistral-7B Family}
        \label{fig:heatmap-mistral}
        \vspace{-1em}
    \end{subfigure}
\end{center}
\caption{
Density ratio rewards from various numerator and denominator model pairings, following Equation (\ref{eq:weak_strong_ratio}). Models, fine-tuned with different objectives, are ordered by their human-aligned levels measured by ArenaHard. Generally, larger alignment gaps between numerator and denominator models yield stronger reward functions, supporting the ``Strong-over-Weak Hypothesis'' in our reward design. This trend holds across models fine-tuned with distinct objectives. An exception, Instruct(PPO)—an official Meta instruct model—achieves a strong ArenaHard score likely due to more intensive SFT training rather than improved human alignment.}
\label{fig:mistral-optimization-rb}
\vspace{-1.5em}
\end{figure*}

\paragraph{Results}

Our findings, visualized in Figure~\ref{fig:dot-plot}, reveal a strong correlation between the accuracy of the reward function in Equation (\ref{eq:weak_strong_ratio}) and the strong-over-weak alignment gap. As the alignment gap widens, the reward function achieves stronger results. When the alignment gap is near zero, the signal becomes noisy, with the RewardBench accuracy approximating 50\%, indicative of a random guess. Further details are presented in Figure~\ref{fig:mistral-optimization-rb}, where each row represents a numerator model and each column a denominator model. Each cell displays the reward function's RewardBench score. The heatmap illustrates that the choice of denominator model significantly impacts reward generalization. Selecting weaker denominator models (e.g., Base or SFT) to ensure a sufficient alignment gap typically results in more effective and stable reward functions. 

The experiment also shows considerable flexibility in constructing density ratio reward. For instance, as shown in Figure~\ref{fig:dot-plot} (left), SFT-RDPO as the numerator performs well with various checkpoints---such as Base, SFT, KTO, RRHF, SLiC-HF, and IPO---as denominators, producing high reward accuracy likely due to these models being less aligned than RDPO. Conversely, using a stronger model as the denominator with SFT-RDPO as the numerator leads to a noticeable drop in reward accuracy. Finally, when Base or SFT models serve as the denominator, nearly any preference-tuned numerator model yields an effective reward function, underscoring that the key to effective reward performance lies in maintaining a meaningful alignment gap rather than requiring DPO or other preference-specific tuning for the numerator model.

\subsection{Customized Strong-Over-Weak Density Ratio}
\label{sec:exp-rb}

\methodname proposes to use customized instructions and in-context learning (ICL) examples to enhance control and accuracy over the vanilla strong-over-weak density ratio. We examine the effect of prompt-based customization in following experiments.

\paragraph{Setup}

We select \textit{Nous-Hermes-2-Mistral-7B-DPO} ~\citep{Nous-Hermes-2-Mistral-7B-DPO} and \textit{OpenHermes-2.5-Mistral-7B} as the model pair in \methodname. 
To tailor vanilla density ratio to specific domains, we develop three customized instruction sets to enhance reward accuracy in Safety, Code/Math, and ChatHard domains. The Safety set focuses on sensitive or high-risk topics like ethics, harmful behavior, profanity, and legal issues, promoting safe and responsible responses. The Code/Math set targets coding tasks and mathematical problem-solving, prioritizing logical reasoning, accuracy, and precision. The ChatHard set emphasizes detailed, nuanced understanding for complex instruction-following tasks. Each set includes domain-specific guidelines and in-context examples (ICLs) showcasing positive and negative cases, enabling the reward function to produce more precise scores. An adaptive router, powered by a zero-shot prompted LLM, assigns the most relevant instruction set to each sample, improving domain adaptability.

\begin{table*}[th]
\centering
\definecolor{lightblue}{rgb}{0.88, 0.92, 1} 
\begin{tabular}{l||c|c|c|c||c}
\toprule
\rowcolor{lightblue} 
\textbf{Reward Function} & \textbf{Chat} & \textbf{ChatHard} & \textbf{Safety} & \textbf{Reasoning} & \textbf{Overall} \\
\midrule
GPT-4-turbo  & 95.3 & 75.4 & 86.7 & 82.7 & 85.2 \\
Claude-3.5-sonnet  & 96.4 & 74.0 & 81.6 & 84.7 & 84.2 \\
RM-Mistral-7B & 96.6 & 60.5 & 87.0 & 77.4 & 80.4 \\
ArmoRM-Llama-3-8B & 96.9 & 76.8 & 90.5 & 97.3 & 90.4 \\
\hline
DPO model-as-a-judge  & 53.0 & 49.5 & 48.3 & 52.1 & 50.0 \\
\hline
density ratio (DPO vs. base) & 89.9 & 65.6 & 62.8 & 71.9 & 71.9 \\
density ratio (SFT vs. base) & 79.6 & 65.6 & 52.8 & 70.0 & 67.0 \\
\rowcolor{lightblue} 
\hline
DPO vs SFT & & & & & \\
vanilla density ratio & 92.2 & 60.5 & 82.4 & 73.8 & 77.2 \\
\methodname (safety) & 88.3 & 61.8 & 91.0 & 87.7 & 82.5 \\
\methodname (code/math) & 91.6 & 60.1 & 89.9 & 89.7 & 83.0 \\
\methodname (chat-hard) & 89.1 & 69.7 & 89.1 & 85.9 & 83.5 \\
\hline
\methodname (adaptive, chat-hard, oracle)  & 89.1 & 69.7 & 91.0 & 89.7 & 84.9 \\
\methodname (adaptive, oracle)  & 92.2 & 60.5 & 91.0 & 89.7 & 83.4 \\
\methodname (adaptive, router)  & 93.9 & 56.8 & 91.0 & 88.0 & 82.6 \\
\bottomrule
\end{tabular}
\caption{Performance on Reward Bench across multiple dimensions (Chat, ChatHard, Safety, and Reasoning). The overall score is the average of these four. RM-Mistral-7B is the strongest in-class trained reward model initialized from \textit{mistralai/Mistral-7B-Instruct-v0.2}. ArmoRM-Llama-3-8B is a SoTA reward model scoring second on RewardBench by time of writing. GPT-4 and Claude-3.5 are proprietary models serving as examples of LLM-as-a-judge reward functions. To construct the density ratio, we can use a DPO model (\textit{Nous-Hermes-2-Mistral-7B-DPO}), an SFT model ( \textit{OpenHermes-2.5-Mistral-7B)}, or a Base model (\textit{Mistral-7B-v0.1}). We denote specific pairings in the format (dpo vs. sft), which, for example, indicates the density ratio between DPO and SFT models. \methodname applies domain-specific instructions (e.g., safety or code/math or chat-hard) when taking density ratio. Adaptive routing configurations include an ``oracle'' (ideal routing) and a real-world ``router'' based on a zero-shot prompted LLM.}
\label{tab:reward_bench_performance}
\vspace{-1.5em}
\end{table*}

\paragraph{Results}

The results in Table~\ref{tab:reward_bench_performance} show a clear benefit of employing \methodname approaches across various dimensions. \methodname reward function is shown to consistently outperform vanilla density ratio without domain-customized instructions. \methodname reward optimized for \textit{safety} achieve a Safety score of 91.0, representing a 7.6-point improvement over uninstructed density ratio baselines. This highlights the benefits of safety-specific guidance in enhancing reward function's safety considerations. Similarly, \methodname tailored for \textit{code/math} achieves a Reasoning score of 89.7, outperforming GPT-4-turbo and Claude-3.5-sonnet, with a substantial 15.9-point gain over baselines. \methodname focused on \textit{chat-hard} scores 69.7 in ChatHard, reflecting improved reward robustness in challenging dialog contexts.

\methodname uses an \textit{oracle} (idealized routing) to establish a performance upper-bound with dynamic routing. Under ideal conditions, it achieves an overall score of 84.9, balancing safety, reasoning, and conversational robustness. In practice, adaptive \methodname employs a \textit{router} (a zero-shot LLM) to automate domain assignment. Notably, the router uses the vanilla density ratio for the general chat domain, as it performs best in Chat, which is the most frequent scenario in real-world annotation settings.

Overall, \methodname outperforms standard density ratio baselines by as much as 5.4 points, showing the advantages of adaptively customized reward functions. Generative reward using the same strong model with an identical instruction set performs near random chance. In contrast, \methodname that contrasts the strong model versus a weaker model achieves 82.6 overall. The performance is comparable to LLM-as-a-judge reward from GPT-4-turbo and Claude-3.5-sonnet, and surpasses the best in-class Mistral-7B classifier reward.

\subsection{Alignment with Density Ratio Annotated Data}
\label{sec:exp-pt}

Previous experiments indicated that \methodname delivers a strong reward signal, achieving high scores on standard reward benchmarks. Here, we preference-tune LLMs using data annotated by \methodname, enabling direct comparisons between \methodname and SoTA reward functions in their effectiveness for preference alignment.

\paragraph{Setup}
We initialize with Meta-Llama-3-8B-Instruct and preference-tune it using SimPO~\citep{Meng2024SimPOSP} with data annotated by \methodname, along with other reward functions (see Appendix~\ref{subsec::annotation} for details). Details about the SimPO algorithm and our training setup are available at Appendix~\ref{sec:simpo_training}. Our evaluation methods include AlpacaEval2.0, ArenaHard, and MT-Bench (details in Appendix~\ref{sec:evaluation}).

\begin{table*}[t]
\centering
\definecolor{lightblue}{rgb}{0.88, 0.92, 1} 
\begin{tabular}{l||ccc||cc||c}
\toprule
\rowcolor{lightblue} 
\textbf{Reward Function} & \multicolumn{3}{c||}{\textbf{AlpacaEval 2}} & \multicolumn{2}{c||}{\textbf{Arena-Hard}} & \textbf{MT-Bench} \\ 
\cline{2-7}
 & LC (\%) & WR (\%) & Length & WR (\%) & Length & GPT-4 \\ 
\midrule
\midrule
N/A (starting model) & 22.9 & 22.6 & 1899 & 22.3 & 596 & 8.1 \\
\hline
ArmoRM-Llama-3-8B & \textbf{55.2} & \textbf{48.2} & 1651 & 30.6 & 475 & 8.0 \\ 
\hline
\rowcolor{lightblue} %
SFT vs Base & & & & & & \\
vanilla density ratio & 23.3 & 21.3 & 1720 & 23.5 & 564 & \textbf{8.3} \\
\methodname (adaptive) & 27.5 & 26.7 & 1888 & 30.4 & 607 & \textbf{8.3} \\
\hline
\rowcolor{lightblue} %
DPO vs SFT & & & & & & \\
vanilla density ratio & 39.9 & 40.1 & 2008 & 34.6 & 571 & 8.1 \\ 
\methodname (safety) & 30.0 & 44.7 & 2850 & \textbf{39.4} & 777 & 8.0 \\
\methodname (code/math) & 36.0 & 33.1 & 1853 & 30.4 & 545 & 8.2 \\
\hline
\methodname (adaptive) & 40.7 & \textbf{46.1} & 2229 & \textbf{37.4} & 643 & 8.0 \\
\bottomrule
\end{tabular}
\caption{Alignment performance after SimPO training on the \textbf{Llama-3-Instruct (8B) model}. Reward function is used to annotate the online preference dataset, obtained through Best-of-32 sampling. The first row is the performance of the starting model \textit{Llama-3-Instruct (8B)} model. The second row is the alignment performance of aligning using a SoTA trained reward function. DPO model indicated is \textit{NousResearch/Nous-Hermes-2-Mistral-7B-DPO}; SFT model is \textit{teknium/OpenHermes-2.5-Mistral-7B}; Base model is \textit{mistralai/Mistral-7B-v0.1}. \methodname applies domain-specific guidance (e.g., safety or code/math) to the vanilla density ratio reward. Adaptive indicates using a routing system to assign domain-related instruction set for each example.}
\label{tab:generative_eval}
\vspace{-1.5em}
\end{table*}

\paragraph{Reward Functions} 
We focus on two model pairs in the \methodname reward formulation: (i) SFT vs. Base, and (ii) DPO vs. SFT. The first model pair (SFT vs. Base) is chosen because neither model has undergone preference tuning, allowing us to test whether a preference reward can be derived based purely on the overall capability improvement after SFT training. The second model pair (DPO vs. SFT) is selected for its reward performance, as shown in Table~\ref{tab:reward_bench_performance}. For the prompt-guided reward function, we experiment with various instruction types: no instructions, safety domain instructions, math/coding domain instructions, and adaptive instructions tailored to the domain of each input prompt.

\paragraph{Results}
As shown in Table~\ref{tab:generative_eval}, Llama-3-instruct preference fine-tuned using data annotated by the DPO-over-SFT density ratio achieve strong performance, with 39.9 on AlpaceEval 2 and 34.6 on ArenaHard. In contrast, SFT-over-Base shows limited improvements after preference alignment. Narrow gap in their human-aligned level results in noisy reward signal that fails to annotate preference data effectively. This demonstrates again that the effectiveness of reward function in (\ref{eq:weak_strong_ratio}) depends on a significant gap in human-value alignment between the numerator and denominator models.

Table~\ref{tab:generative_eval} shows that reward functions customized for specific domain can not be applied universally to all examples, doing so would result in suboptimal performance, as in ``safety'' and ``code/math'' \methodname results. We find that by using adaptive instructions---currently categorized into Chat, Code/Math, and Safety--- that finds best specialized reward for each example, we achieve the highest overall alignment performance, with 40.7 on AlpacaEval 2 and 37.4 on ArenaHard, competitive against SoTA reward from ArmoRM. Notably, for the (SFT, base) model pair, adaptive customization of reward significantly enhances alignment performance across all three benchmarks, making a weak density ratio reward signal much more effective.

\section{Related Works}
\paragraph{Preference tuning}
Many preference tuning algorithms have been proposed to align LLMs with human preferences and values \citep{Melnyk2024DistributionalPA, Pang2024IterativeRP, Ethayarajh2024KTOMA, Wu2024SelfPlayPO, Hong2024ORPOMP, Yuan2023RRHFRR}. The most well-known one is the proximal policy optimization \citep[PPO;][]{Schulman2017ProximalPO}, an online RL algorithm that optimizes policy to maximize the KL-constrained reward expectation of an external reward model. Direct preference optimization \citep[DPO;][]{Rafailov2023DirectPO} leverages DPO implicit reward -- parameterized as density ratio between policy model and a reference model---to circumvent the need of external reward function. It simultaneously optimizes the implicit reward and policy model by training on pairwise preference data. More recently, SimPO~\citep{Meng2024SimPOSP} directly optimizes the average log-likelihood margin between winning and losing sequences, eliminating the need for a reference model.

\paragraph{Density ratio reward functions} 
Density ratio as reward function is popularized by implicit DPO reward ~\citep{Rafailov2023DirectPO}. ~\citet{Chen2024BootstrappingLM} uses implicit DPO reward to bootstrap an LLM through iterative DPO training. \citet{Zhong2024DPOMP} trains a DPO model and uses the density ratio to derive a token-level characterization for response quality, and uses it as a reward signal in PPO training. \citet{Yang2024NotAP} uses the density ratio between DPO vs SFT model as quality filter. Though one study~\citet{Lin2024OnTL}  finds that implicit DPO reward struggles to generalize on OOD examples compared with just training a classifier using \citep[BradleyTerry;][]{Bradley1952RankAO} objective. This work extends the density ratio reward formulation to broader spectrum of models, and provides guidance for finding stronger reward signal than implicit DPO reward.

\paragraph{Discriminative \& generative rewards}
Trained classifiers and generative rewards are the mainstream method for preference data annotation. 
They top leaderboards such as RewardBench ~\citep{Lambert2024RewardBenchER} and are widely used to preference align well-known models~\citep{Ouyang2022TrainingLM, Touvron2023Llama2O, Adler2024Nemotron43T, Yang2024Qwen2TR, Cui2023UltraFeedbackBL}. 
High quality and popular preference datasets are often annotated using powerful proprietary models as-a-judge, either in the forms of scalar score or textual assessment and critiques~\citep{Cui2023UltraFeedbackBL}. Then, one can use the data to finetune a generative judge~\citep{Wang2024SelfTaughtE, Zhang2024GenerativeVR, Wang2024DirectJP,kim2024prometheusopen} or to train a sequence classifier~\citep{Adler2024Nemotron43T, Dong2024RLHFWF, skyworkreward2024}. \methodname provides a data-free and training-free alternative for reward modeling and preference annotation.

\paragraph{Weak-to-strong generalization} 
Prior works have explored the idea of contrasting a weak and a strong model to obtain better performance than the strong model. Contrastive decoding (CD), for instance, enhances LLM generation quality by searching for sequences that maximizes the likelihood difference between an expert model and an amateur model. \citet{OBrien2023ContrastiveDI} shows CD consistently improves reasoning tasks. ~\citet{Li2022ContrastiveDO} shows improved generation quality in wikipedia, news and story domains. ~\citet{Chuang2023DoLaDB} shows improvement in LLM facutuality by contrasting the differences between logits in later layers and earlier layers. \textsc{ExPo}~\citep{Zheng2024WeaktoStrongEE} uses the delta between an aligned model and pre-aligned model to extrapolate a better aligned models through weight merging. \methodname similarly contrasts strong-over-weak models, and uses the delta to align small models to near GPT-4 level performance on ArenaHard (Figure~\ref{fig:arena_hard leaderboard}).

\section{Conclusion and Future Work}
We introduce \methodname, a cost-effective and accessible approach that uses off-the-shelf LLMs for preference data annotation. It reduces the need for costly human labeling or proprietary models to achieve a high-performance reward function. At the core of \methodname is the Strong-over-Weak hypothesis, which we rigorously validate through extensive experiments. This insight offers a design guideline for practitioners seeking LLM-based preference annotation.

Domain-specific customization further enhances the density ratio reward, particularly in targeted areas such as safety and reasoning. And this is achieved without requiring additional data or fine-tuning. We offer an automated pipeline to adaptively combine domain-expert reward functions for tailored preference annotation. This approach shows strong performance on reward benchmarks, and its annotated data pushes an 8B model to GPT-4 level performance on ArenaHard (Figure~\ref{fig:arena_hard leaderboard}). This result is competitive with state-of-the-art (SoTA) reward classifiers while avoids the data and compute overheads of actually training reward functions, highlighting \methodname as both cost-effective and highly effective.

Recently, density ratio based reward functions have demonstrated state-of-the-art performance as Math Process-Reward Models (PRMs)~\citep{yuan2024implicitprm}, as it provides token-level value estimates. Exploring the use of \methodname for process-level presents a promising future direction, particularly for inference-time scaling use-cases.

\nocite{langley00}

\clearpage
\bibliography{referece}
\bibliographystyle{icml2025}

\clearpage
\appendix

\onecolumn
\section{Experimental Details}

\subsection{Preference Data Annotation}
\label{subsec::annotation}

We use input prompts $\mathcal{D} = \{x^{(i)}\}_{i=1}^N$ from the UltraFeedback dataset~\citep{Cui2023UltraFeedbackBL}. On-policy alignment dataset is created by Best-of-N sampling, and constructing chosen/rejected pairs using different reward functions. 
For each prompt $x \in \mathcal{D}$, we sample 32 model completions $\{y_i\}_{i=1}^{32}$ from the starting policy. To construct positive-negative paired preference data, we select the preferred response $y_{i^\ast}$ as the one that maximizes the reward function: $i^\ast = \arg\max_i r(x, y_i)$. A dispreferred response is then randomly sampled  from the remaining set. For all experiments, the completions $\{y_i\}_{i=1}^{32}$ are pre-computed and fixed, with only the choice of reward function $r$ varying, as indicated in the Reward Function column in Table~\ref{tab:generative_eval}. To address possible length imbalances between preferred and dispreferred responses, we apply a length threshold before randomly selecting the rejected sample. This procedure ensures variety in rejected samples, reduces the risk of reward hacking, and maintains a length-balanced preference dataset.

\subsection{Training Details}
\label{sec:simpo_training}

\paragraph{Training Details}
We use SimPO \citep{Meng2024SimPOSP} as our preference optimization method, which optimizes the average log-likelihood margin between positive and negative responses directly without requiring a reference model. Its loss function is:
\begin{equation}
    -\log \sigma\left({\beta \over \lVert y_{\text{accept}} \rVert} \log \pi(y_{\text{accept}} \mid x) - {\beta \over \lVert y_{\text{reject}} \rVert} \log \pi(y_{\text{reject}} \mid x) - \gamma \right),
\end{equation}

where $\sigma$ is the sigmoid function, $\beta$ is the scaling term for reward difference, and $\gamma$ is the reward margin term. 
We choose SimPO for its strong alignment results, matching or even outperforming those of DPO, with the added advantage of better efficiency by eliminating the memory and compute demands of a reference model.

To account for SimPO's training instability and ensure fair comparison of reward functions, we perform hyper-parameter search for each preference dataset. We explore the following hyper-parameters ranges: learning rate in [5e-7, 8e-7  1e-6] and $\beta$ in [10.0, 18.0]. We fix the $\gamma$ / $\beta$ ratio to be 0.3 since our experiments show that it has limited effect on final model performance. A batch size of 128 and one training epoch are used for all experiments according to the initial setup in \citet{Meng2024SimPOSP}. Additionally, we set the max sequence length to 2048 and apply a cosine learning rate scheduler with 10\% warm-up steps.

\section{Evaluation}
\label{sec:evaluation}
\paragraph{RewardBench} We use RewardBench~\citep{Lambert2024RewardBenchER} to evaluate DR's out-of-distribution reward performance. It is a comprehensive benchmark designed test the performance of reward models across a range of scenarios, including challenging, clean, and out-of-distribution (OOD) tasks. The dataset consists of 2,850 prompt-chosen-rejected trios, where reward models are tasked with accurately identifying the preferred response. RewardBench is structured around four key dimensions—Chat, ChatHard, Safety, and Reasoning—each targeting different capabilities of the models. The overall RewardBench score is calculated by averaging the classification accuracy across these dimensions, providing a balanced assessment of model performance.

\paragraph{ArenaHard} We use ArenaHard~\citep{Li2024FromCD} score as proxy for a model's human preferred level, it is shown to have the highest correlation and separability against gold human judgments in ChatArena. While it doesn't not score individual dimensions of preference, it provides an aggregate signal for overall human preference. The delta is calculated as the difference between strong model and weak model's arena hard score.

\paragraph{AlpacaEval2.0} Both AlapcaEval2.0~\citep{Dubois2024LengthControlledAA} and ArenaHard are win-rate based metrics against answers generated by a reference model; and we use the recommended default choices of reference models and judge models for both benchmarks.  AlpacaEval2.0 addresses LLM-as-a-judge's bias for longer responses by providing a length adjusted win-rate that better correlates with human ranking.

\paragraph{MT-Bench}  MT-Bench~\citep{Zheng2023JudgingLW} is a multi-turn benchmark that measures model performance on 8 dimensions compared to a reference ground-truth.

\section{Models Used for Density Ratio Reward Experiments}
\subsection{Iterative DPO Models}
The checkpoints for our experiment on density ratio reward for iterative DPO checkpoints in Figure~\ref{fig:iter-dpo} are off-the-shelf models released by \citet{Meng2024SimPOSP} and \citet{Chen2024BootstrappingLM}. Details are summarized in the following tables.

\begin{table*}[ht!]
\centering
\definecolor{lightblue}{rgb}{0.88, 0.92, 1} 
\begin{tabular}{|l|l|c|}
\hline
\rowcolor{lightblue} 
\textbf{PaperName} & \textbf{HuggingfaceModel} & \textbf{ArenaHard} \\
\hline
Base      & \href{https://huggingface.co/mistralai/Mistral-7B-v0.1}{mistralai/Mistral-7B-v0.1}           & 0         \\
SFT       & \href{https://huggingface.co/alignment-handbook/zephyr-7b-sft-full}{alignment-handbook/zephyr-7b-sft-full} & 6.2       \\
DPO-iter0 & \href{https://huggingface.co/HuggingFaceH4/zephyr-7b-beta}{HuggingFaceH4/zephyr-7b-beta}     & 12.5      \\
DPO-iter1 & \href{https://huggingface.co/sail/Zephyr-7B-DICE-Iter1}{sail/Zephyr-7B-DICE-Iter1}           & 15.5      \\
DPO-iter2 & \href{https://huggingface.co/sail/Zephyr-7B-DICE-Iter2}{sail/Zephyr-7B-DICE-Iter2}           & 16.7      \\
\hline
\end{tabular}
\caption{Mistral Iterative DPO Checkpoints}
\label{tab:mistral_checkpoint_source}
\end{table*}

\begin{table*}[ht!]
\centering
\definecolor{lightblue}{rgb}{0.88, 0.92, 1} 
\begin{tabular}{|l|l|c|}
\hline
\rowcolor{lightblue} 
\textbf{PaperName} & \textbf{HuggingfaceModel} & \textbf{ArenaHard} \\
\hline
Base      & \href{https://huggingface.co/meta-llama/Meta-Llama-3-8B}{meta-llama/Meta-Llama-3-8B}                       & 0         \\
SFT       & \href{https://huggingface.co/princeton-nlp/Llama-3-Base-8B-SFT}{princeton-nlp/Llama-3-Base-8B-SFT}         & 3.1       \\
DPO-iter0 & \href{https://huggingface.co/princeton-nlp/Llama-3-Base-8B-SFT-DPO}{princeton-nlp/Llama-3-Base-8B-SFT-DPO} & 15.9      \\
DPO-iter1 & \href{https://huggingface.co/sail/Llama-3-Base-8B-DICE-Iter1}{sail/Llama-3-Base-8B-DICE-Iter1}             & 35.9      \\
DPO-iter2 & \href{https://huggingface.co/sail/Llama-3-Base-8B-DICE-Iter2}{sail/Llama-3-Base-8B-DICE-Iter2}             & 41.2      \\
\hline
\end{tabular}
\caption{Llama Iterative DPO Checkpoints}
\label{tab:llama_checkpoint_source}
\end{table*}

\subsection{Models Trained via Diverse Preference Optimization Objectives}
The checkpoints for experiment in Section~\ref{sec:exp-weak-to-strong} are taken from existing works~\citep{Meng2024SimPOSP} with details listed below.

\begin{table*}[ht!]
\definecolor{lightblue}{rgb}{0.88, 0.92, 1} 
\centering
\begin{tabular}{|l|l|c|c|}
\hline
\rowcolor{lightblue} 
\textbf{PaperName} & \textbf{HuggingfaceModel} & \textbf{AlpacaEval2.0} & \textbf{ArenaHard} \\
\hline
Base            & \href{https://huggingface.co/mistralai/Mistral-7B-v0.1}{mistralai/Mistral-7B-v0.1}                             & 0.0  & 0.0   \\
SFT             & \href{https://huggingface.co/alignment-handbook/zephyr-7b-sft-full}{alignment-handbook/zephyr-7b-sft-full}       & 8.4  & 1.3   \\
SFT-CPO         & \href{https://huggingface.co/princeton-nlp/Mistral-7B-Base-SFT-CPO}{princeton-nlp/Mistral-7B-Base-SFT-CPO}       & 9.8  & 6.9   \\
SFT-KTO         & \href{https://huggingface.co/princeton-nlp/Mistral-7B-Base-SFT-KTO}{princeton-nlp/Mistral-7B-Base-SFT-KTO}       & 13.1 & 5.6   \\
SFT-DPO         & \href{https://huggingface.co/princeton-nlp/Mistral-7B-Base-SFT-DPO}{princeton-nlp/Mistral-7B-Base-SFT-DPO}       & 15.1 & 10.4  \\
SFT-RDPO        & \href{https://huggingface.co/princeton-nlp/Mistral-7B-Base-SFT-RDPO}{princeton-nlp/Mistral-7B-Base-SFT-RDPO}     & 17.4 & 9.9   \\
SFT-IPO         & \href{https://huggingface.co/princeton-nlp/Mistral-7B-Base-SFT-IPO}{princeton-nlp/Mistral-7B-Base-SFT-IPO}       & 11.8 & 7.5   \\
SFT-SLiC-HF     & \href{https://huggingface.co/princeton-nlp/Mistral-7B-Base-SFT-SLiC-HF}{princeton-nlp/Mistral-7B-Base-SFT-SLiC-HF} & 10.9 & 7.3   \\
SFT-RRHF        & \href{https://huggingface.co/princeton-nlp/Mistral-7B-Base-SFT-RRHF}{princeton-nlp/Mistral-7B-Base-SFT-RRHF}     & 11.6 & 6.9   \\
SFT-SimPO       & \href{https://huggingface.co/princeton-nlp/Mistral-7B-Base-SFT-SimPO}{princeton-nlp/Mistral-7B-Base-SFT-SimPO}   & 21.4 & 16.6  \\
SFT-ORPO        & \href{https://huggingface.co/kaist-ai/mistral-orpo-beta}{kaist-ai/mistral-orpo-beta}                             & 14.7 & 7.0   \\
\hline
\end{tabular}
\caption{Mistral Models trained with various preference optimization objectives; checkpoints used for our Strong-over-Weak experiments in Section~\ref{sec:exp-weak-to-strong} }
\label{tab:mistral_simpo_checkpoints}
\end{table*}

\begin{table*}[ht!]
\definecolor{lightblue}{rgb}{0.88, 0.92, 1} 
\centering
\begin{tabular}{|l|l|c|c|}
\hline
\rowcolor{lightblue} 
\textbf{PaperName} & \textbf{HuggingfaceModel} & \textbf{AlpacaEval2.0} & \textbf{ArenaHard} \\
\hline
Base                  & \href{https://huggingface.co/meta-llama/Meta-Llama-3-8B}{meta-llama/Meta-Llama-3-8B}                             & 0.0   & 0.0    \\
SFT                              & \href{https://huggingface.co/princeton-nlp/Llama-3-Base-8B-SFT}{princeton-nlp/Llama-3-Base-8B-SFT}               & 6.2   & 3.3    \\
SFT-CPO                          & \href{https://huggingface.co/princeton-nlp/Llama-3-Base-8B-SFT-CPO}{princeton-nlp/Llama-3-Base-8B-SFT-CPO}       & 10.8  & 5.8    \\
SFT-ORPO                         & \href{https://huggingface.co/princeton-nlp/Llama-3-Base-8B-SFT-ORPO}{princeton-nlp/Llama-3-Base-8B-SFT-ORPO}     & 12.2  & 10.8   \\
SFT-KTO                          & \href{https://huggingface.co/princeton-nlp/Llama-3-Base-8B-SFT-KTO}{princeton-nlp/Llama-3-Base-8B-SFT-KTO}       & 14.2  & 12.5   \\
SFT-DPO                          & \href{https://huggingface.co/princeton-nlp/Llama-3-Base-8B-SFT-DPO}{princeton-nlp/Llama-3-Base-8B-SFT-DPO}       & 18.2  & 15.9   \\
SFT-RDPO                         & \href{https://huggingface.co/princeton-nlp/Llama-3-Base-8B-SFT-RDPO}{princeton-nlp/Llama-3-Base-8B-SFT-RDPO}     & 17.6  & 17.2   \\
SFT-IPO                          & \href{https://huggingface.co/princeton-nlp/Llama-3-Base-8B-SFT-IPO}{princeton-nlp/Llama-3-Base-8B-SFT-IPO}       & 14.4  & 17.8   \\
SFT-SimPO                        & \href{https://huggingface.co/princeton-nlp/Llama-3-Base-8B-SFT-SimPO}{princeton-nlp/Llama-3-Base-8B-SFT-SimPO}   & 22.0  & 20.3   \\
Instruct (PPO)                         & \href{https://huggingface.co/meta-llama/Meta-Llama-3-8B-Instruct}{meta-llama/Meta-Llama-3-8B-Instruct}           & 26.0  & 22.3   \\
\hline
\end{tabular}
\caption{Llama Model Comparison with AlpacaEval2.0 and ArenaHard Scores}
\label{tab:llama_simpo_checkpoints}
\end{table*}

\begin{figure*}[h]
    \centering
    \begin{tcolorbox}
        \small
        \texttt{<s><|im\_start|>system} \\
        \texttt{\{system\_prompt\}} \\
        \texttt{<|im\_end|>} \\
 
        \texttt{<|im\_start|>user} \\
        \texttt{You must carefully understand my question and give a relevant, correct, and logical answer.}

        \vspace{0.5em}
        \texttt{For example:} \\
        \texttt{User: \{ICL\_query\_i\}} \\
        \texttt{Good Assistant: \{chosen\_response\_i\}} \\
        \texttt{Bad Assistant: \{rejected\_response\_i\}} \\
        \texttt{Explanation: \{explanation\_i\}}

        \vspace{0.5em}
        \texttt{User: \{user\_query\}} \\
        \texttt{Good Assistant: } \\
        \texttt{<|im\_end|>}
    \end{tcolorbox}
    \caption{Few-shot Instruction template to guide rewards.}
    \label{fig:prompt-template}
\end{figure*}

\begin{figure*}[h]
    \centering
    \begin{tcolorbox}[colback=yellow!5!white, colframe=yellow!75!black, width=\textwidth]
        \small
        
        You are a helpful AI assistant. \\[1em]
        
        You should avoid providing information or responding to questions that could be used to spread hate, incite violence, or promote harmful stereotypes, and instead redirect or rephrase the question to encourage education, understanding, and empathy. \\[1em]
        
        You should prioritize promoting respectful, inclusive, and evidence-based dialogue while mitigating potential harm by rephrasing or redirecting questions that may be misinformed, sensational, or divisive, and by providing accessible information that fosters critical thinking and empathy, especially when handling queries that touch on sensitive topics, explore the gray areas of science, or express potential harm to individuals or communities.
    \end{tcolorbox}
    \caption{Safety guidelines generated by Llama-3.1-8B-Instruct. The prompts are automatically tuned from PKU-Alignment/PKU-SafeRLHF dataset.}
    \label{fig:ai_guidelines}
\end{figure*}

\begin{figure*}[h]
    \centering
    \begin{tcolorbox}[colback=yellow!5!white, colframe=yellow!75!black, width=\textwidth]
        \small
        \textbf{Guideline 1: Answer Constructive, Clear Questions} \\
        Provide an answer when the user asks for factual information, constructive advice, or help with personal growth. Focus on offering practical, positive guidance.
    \end{tcolorbox}
    
    \vspace{0.5em}
    
    \begin{tcolorbox}[colback=yellow!5!white, colframe=yellow!75!black, width=\textwidth]
        \small
        \textbf{Guideline 2: Recognize Jokes, Puns, and Fictional Contexts} \\
        Respond playfully when the question references humor, games, movies, or fictional scenarios. Acknowledge the fictional nature while keeping the tone light.
    \end{tcolorbox}
    
    \vspace{0.5em}
    
    \begin{tcolorbox}[colback=yellow!5!white, colframe=yellow!75!black, width=\textwidth]
        \small
        \textbf{Guideline 3: Avoid Answering Harmful, Illegal, or Malicious Questions} \\
        Do not engage if the question promotes harm, illegal activities, or unethical behavior. Politely but firmly refuse to provide an answer, while keeping the response respectful.
    \end{tcolorbox}
    
    \vspace{0.5em}
    
    \begin{tcolorbox}[colback=yellow!5!white, colframe=yellow!75!black, width=\textwidth]
        \small
        \textbf{Guideline 4: Handle Sensitive Topics with Empathy} \\
        Respond with care to questions about mental health, personal relationships, or emotionally charged situations. Acknowledge the user's feelings, and offer general advice or suggest professional resources.
    \end{tcolorbox}
    
    \vspace{0.5em}
    
    \begin{tcolorbox}[colback=yellow!5!white, colframe=yellow!75!black, width=\textwidth]
        \small
        \textbf{Guideline 5: Clarify Ambiguous or Potentially Problematic Questions} \\
        Ask for clarification if a question seems unclear, possibly problematic, or if it could have multiple interpretations. Avoid jumping to conclusions, and invite the user to explain further.
    \end{tcolorbox}
    
    \caption{The five safety guidelines used for the ablation study. Guidelines 1-4 were adopted in the final system, while Guideline 5 was excluded due to performance regression.}
    \label{fig:safety_guidelines}
\end{figure*}

\section{Ablation on Prompt Design}

We started our prompt experiment with a simple \textbf{seed} prompt: \emph{``You are a helpful AI assistant.''}, we surprising observe an improvement of 2.9 points on the RewardBench score. This result is unexpected, as it demonstrates that even minimal prompting can significantly enhance performance. Notably, most of the gains occur in the \textbf{Reasoning} domain in RewardBench, which covers coding and math domains.

To better understand the performance gains from applying instructions to density ratio, we ablate the effect of incrementally adding Safety Instructino in Figure~\ref{fig:safety_guidelines}. The results are shown in Table~\ref{tab:prompt_performance}, where safe1 adds the first safety principle to the seed prompt, safe2 adds the second principle on safe1, and so on so forth.

\begin{itemize}[leftmargin=1.5em]
    \item \textbf{safe1} includes only the first safety guideline.
    \item \textbf{safe2} incorporates the first two guidelines.
    \item \textbf{safe3} builds on this with three guidelines.
    \item \textbf{safe4}, our final design, includes all four safety guidelines.
    \item \textbf{safe5}, adds additional guideline, but leads to performance regression.
\end{itemize}

Interestingly, while adding the first few guidelines (\textbf{safe1} to \textbf{safe3}) yielded consistent improvements in \textbf{Safety} scores, up until the fourth guideline (\textbf{safe4}) shows diminishing returns and even slight regressions in some domains like Reasoning. Adding the fifth guideline (\textbf{safe5}) led to performance degradation, suggesting that overloading the prompt with rules may reduce effectiveness. Ultimately, we selected \textbf{safe4} as our final configuration, as it provides comprehensive coverage of safety scenarios while balancing performance across domains. However, we also find that leaner prompts like \textbf{safe2} or \textbf{safe3} deliver comparable results in safety-focused metrics. In the last two rows, we report the complete \methodname setup combining guidelines and ICL examples, where the performance gains become more significant.

\begin{table*}[ht]
\centering
\definecolor{lightblue}{rgb}{0.88, 0.92, 1} 
\definecolor{lightgreen}{rgb}{0.88, 0.97, 0.88} 
\begin{tabular}{l||c|c|c|c||c}
    \toprule
    \rowcolor{lightblue} 
    \textbf{Prompt} & \textbf{Chat} & \textbf{ChatHard} & \textbf{Safety} & \textbf{Reasoning} & \textbf{Overall} \\
    \midrule
    -          & 92.2 & 60.5 & 82.4 & 73.8 & 77.2 \\
    \hline
    seed          &  91.1    &  60.8    &  83.5    &  87.8    &  80.1    \\
    \hline
    safe1           & 93.8 & 56.8 & 83.9 & 81.2 & 79.0 \\
    safe2           & 94.7 & 57.7 & 89.3 & 82.6 & 81.1 \\
    safe3           & 93.0 & 60.1 & 90.2 & 82.4 & 81.7 \\
    \rowcolor{lightblue}
    safe4-final         & 91.1 & 59.2 & 91.6 & 77.6 & 79.9 \\
    safe5         & 89.4 & 55.9 & 87.8 & 74.9 & 77.0 \\
    \rowcolor{lightgreen}
    auto-safe     & 91.9 & 53.7 & 87.6 & 80.0 & 78.3 \\
    \hline
    \rowcolor{lightblue}
    safe4-final + ICL   & 88.3 & 61.8 & 91.0 & 87.7 & 82.5 \\
    \rowcolor{lightgreen}
    auto-safe + ICL  & 90.2  & 60.3  & 89.8  & 86.9  & 81.8  \\
    \bottomrule
    \end{tabular}
    \caption{RewardBench Performance ablating the rules and criterion to arrive at our final Safety system prompt -- safe4; light-green highlights an automatically generated safety prompt, auto-safe, which is tuned on the PKU-Alignment/PKU-SafeRLHF ~\citep{Ji2024PKUSafeRLHFTM}. We find the automatic prompt generalizes well to the held-out RewardBench evaluation, giving competitive performance to human-written prompts.}
    \label{tab:prompt_performance}
\end{table*}

\begin{table*}[h]
    \centering
    \definecolor{lightblue}{rgb}{0.88, 0.92, 1} 
    \begin{tabular}{@{}lcccccc@{}}
        \toprule
        \rowcolor{lightblue}
        \textbf{ICL-example} & \textbf{Chat} & \textbf{ChatHard} & \textbf{Safety} & \textbf{Reasoning} & \textbf{Overall} \\ 
        \midrule
        -          & 92.2 & 60.5 & 82.4 & 73.8 & 77.2 \\
        \midrule
        \rowcolor{lightblue}
        \multicolumn{6}{l}{\textbf{sys. + ChatHard ICLs}} \\
        ChatHard1   & 91.1  & 69.1  & 88.0  & 85.9  & 83.5  \\
        ChatHard2   & 93.0  & 63.6  & 88.7  & 88.2  & 83.4  \\
        ChatHard3  & 88.8  & 69.3  & 88.7  & 87.2  & 83.5  \\
        ChatHard4  & 89.9  & 66.0  & 91.9  & 86.6  & 83.6  \\
        ChatHard5  & 90.5  & 63.8  & 91.7  & 89.7  & 83.9  \\
        ChatHard6  & 94.7  & 59.9  & 89.2  & 89.3  & 83.4  \\

        \rowcolor{lightblue}
        \multicolumn{6}{l}{\textbf{sys. + Safety ICLs}} \\
        Safe1      & 88.3  & 61.8  & 91.0  & 87.9  & 82.3  \\
        Safe2      & 90.8  & 64.3  & 89.7  & 86.2  & 82.8  \\
        Safe3      & 91.3  & 60.1  & 91.1  & 87.8  & 82.6  \\

        \rowcolor{lightblue}
        \multicolumn{6}{l}{\textbf{sys. + Math/Coding/Reasoning ICLs}} \\
        Reasoning1 & 91.9  & 59.9  & 90.1  & 88.7  & 82.7  \\
        Reasoning2 & 91.9  & 61.2  & 88.2  & 87.0  & 82.1  \\
        Reasoning3 & 90.2  & 64.3  & 90.0  & 85.8  & 82.6  \\
        Reasoning4 & 90.5  & 61.8  & 89.5  & 88.7  & 82.6  \\
        Reasoning5 & 93.6  & 61.6  & 88.7  & 87.1  & 82.8  \\
        Reasoning6 & 91.6  & 58.8  & 88.8  & 87.5  & 81.7  \\
        Reasoning7 & 88.27 & 60.1  & 89.9  & 87.0  & 81.8  \\
        Reasoning8 & 91.6  & 61.0  & 89.9  & 89.7  & 83.1  \\

        \bottomrule
    \end{tabular}
    \caption{Ablate in-context-learning example's effect on reward performance.}
    \label{tab:domain_icl_results}
\end{table*}

\subsection{Automatic Prompt Tuning for Target Domains}\label{app:auto-prompt}
While reward customization through prompting is effective and does not require fine-tuning, finding a set of preference instructions that works well for your target domain may be challenging. We take inspiration from automatic prompt search/tuning literature~\citep{DOosterlinck2024InContextLF}, and implement an automatic prompt tuning algorithm for a target domain. 

The algorithm goes as follows:\\
Given an initial seed prompt $S$, domain dataset $D$ containing (chosen, rejected) pairs, and an accuracy-metric $Metric(p)$, we iteratively refine the prompt to maximize the accuracy metric on the target domain dataset. The metric is simply ~\methodname's accuracy on the domain dataset.
Let $\text{current\_prompt} = S$ initially. At each iteration $i$, we generate $N$ candidate guidelines using a large language model (We use Llama-3.1-8B-Instruct). For each candidate instruction $c$, we evaluate $Metric(\text{current\_prompt} + c)$ . If the best candidate improves the current reward, we update $\text{current\_prompt}$ accordingly. This process continues for a maximum number of iterations or until no improvement is found, returning the optimized prompt.

The key advantage of this approach is its ability to automatically explore the prompt space guided by a metric $Metric(p)$. The method requires only: (1) an initial prompt, (2) a quality metric, and (3) domain-wise data for evaluation purpose, making it broadly applicable across domains.

We used the above described algorithm to automatically generate instructions for the safety domain. The LLM used to generate prompt is Llama-3.1-8B-Instruct, and we used PKU-SafeRLHF as the domain dataset to evaluate instruction quality. The resulting prompt (Figure~\ref{fig:ai_guidelines}) give comparable performance to human crafted prommpts as shown in Table~\ref{tab:prompt_performance}.

\subsection{Domain-specific In-context Examples}
We created a pool of demonstrations or in-context learning (ICL) examples and grouped them by their primary intended domains, such as ChatHard, Safety, and Reasoning(Math/Code). Although some ICL examples span multiple domains—for instance, the reasoning example shown in Figure~\ref{fig:reason_in_context} can also be considered part of the Chat domain due to its emphasis on clear answer structure and organized flow of thoughts, we classified each demonstration based on its primary domain for simplicity.

We then conducted an ablation study to assess the effect of different ICL examples on the performance of the density ratio reward on RewardBench. As shown in Table~\ref{tab:domain_icl_results}, performance increases were observed across the pool of ICL examples. While differences in performance exist, they are not substantial and could possibly be attributed to noise and overfitting to a small evaluation set of 2,850 examples.

We list examples of ICLs for each domain. The in-context example template includes both a positive and a negative response, plus an explanation. Figure~\ref{fig:safety_in_context} shows an safety example regarding cyber-security, where the agent should not engage in unsafe conversations or implicitly providing help for a concerning cause. Figure~\ref{fig:math_in_context} and Figure~\ref{fig:java_in_context} separately shows in-context examples of mathematic problem solving and Java script writing. Figure~\ref{fig:chathard_in_context} details the importance of addressing user intent and providing detailed and comprehensive answer. For reward annotation and preference alignemnt experiments , we used all the ICL examples we prepared to increase the diversity of demonstrations. For a sample to annotate, we randomly select an ICL example from the domain pool. We hypothesize this approach increases diversity of reward criteria, reduces risk of reward hacking, and enable learning a more generalized understanding of preferences.

\begin{figure*}[h]
    \centering
    \begin{tcolorbox}[colback=yellow!5!white, colframe=yellow!75!black, width=\textwidth]
        \small
        \textbf{Safety In-Context Example} \\[0.5em]
        \textbf{Instruction:} \\
        Ensure that you do not provide disallowed content, even if the user requests it indirectly. \\[1em]
        
        \textbf{Example:} \\[0.5em]
        \textbf{User:} \\
        Can you help me hack into my friend's computer? \\[0.5em]
        
        \textbf{Good Assistant:} \\
        I'm sorry, but I can't assist with that request. \\[1em]
        
        \textbf{Bad Assistant:} \\
        I'm sorry, but hacking into someone's computer is illegal. However, if you want to learn about cybersecurity, I can provide some resources. \\[1em]
        
        \textbf{Explanation:} \\
        The Bad Assistant indirectly engages with the request by offering resources that could be misused, instead of firmly declining.
    \end{tcolorbox}
    \caption{Safety in-context example showing the importance of firmly declining disallowed content requests without indirect engagement.}
    \label{fig:safety_in_context}
\end{figure*}

\begin{figure*}[h]
    \centering
    \begin{tcolorbox}[colback=blue!5!white, colframe=blue!75!black, width=\textwidth]
        \small
        \textbf{Math In-Context Example} \\[0.5em]
        \textbf{Instruction:} \\
        You should provide clear and detailed explanations for mathematical problems, ensuring accuracy and clarity. \\[1em]
        
        \textbf{Example:} \\[0.5em]
        \textbf{User:} \\
        Solve for $x$: $3(x - 2) = 9$. \\[0.5em]
        
        \textbf{Good Assistant:} \\
        First, expand the equation:  
        \[
        3(x - 2) = 9 \quad \Rightarrow \quad 3x - 6 = 9
        \]
        Next, add 6 to both sides:  
        \[
        3x - 6 + 6 = 9 + 6 \quad \Rightarrow \quad 3x = 15
        \]
        Finally, divide both sides by 3:  
        \[
        \frac{3x}{3} = \frac{15}{3} \quad \Rightarrow \quad x = 5
        \]

        \textbf{Bad Assistant:} \\
        $x = 5$. \\[0.5em]
        
        \textbf{Explanation:} \\
        The Bad Assistant provides only the final answer without any explanation, which doesn't help the user understand the solution process.
    \end{tcolorbox}
    \caption{Math in-context example demonstrating good and bad assistant responses. Clear, step-by-step explanations are essential for helping users understand mathematical solutions.}
    \label{fig:math_in_context}
\end{figure*}

\begin{figure*}[h]
    \centering
    \begin{tcolorbox}[colback=blue!5!white, colframe=blue!75!black, width=\textwidth]
        \small
        \textbf{Reason In-Context Example} \\[0.5em]
        \textbf{Instruction:} \\
        Please provide responses that are well-structured, using proper grammar and punctuation, and avoid run-on sentences or disorganized thoughts. \\[1em]
        
        \textbf{Example:} \\[0.5em]
        \textbf{User:} \\
        Describe the process of cell division during mitosis. \\[0.5em]
        
        \textbf{Good Assistant:} \\
        Mitosis is a cell division process that results in two genetically identical daughter cells. It consists of stages: prophase, metaphase, anaphase, and telophase, during which the nucleus and genetic material divide. \\[1em]
        
        \textbf{Bad Assistant:} \\
        Cell division in mitosis is when a cell divides into two cells and the chromosomes separate and it goes through prophase metaphase anaphase and telophase. \\[1em]
        
        \textbf{Explanation:} \\
        The Bad Assistant's response lacks proper structure and punctuation, making it a run-on sentence that is difficult to follow.
    \end{tcolorbox}
    \caption{Reason in-context example demonstrating the importance of clear, structured, and grammatically correct responses. }
    \label{fig:reason_in_context}
\end{figure*}

\begin{figure*}[h]
    \centering
    \begin{tcolorbox}[colback=blue!5!white, colframe=blue!75!black, width=\textwidth]
        \small
        \textbf{Java In-Context Example} \\[0.5em]
        \textbf{Instruction:} \\
        Provide clear and helpful explanations along with your code to aid the user's understanding. \\[1em]
        
        \textbf{Example:} \\[0.5em]
        \textbf{User:} \\
        How do I reverse a linked list in Java? \\[0.5em]
        
        \textbf{Good Assistant:} \\
        Here's how you can reverse a singly linked list in Java:
        \begin{lstlisting}
class Node {
    int data;
    Node next;
}

public Node reverseList(Node head) {
    Node prev = null;
    Node current = head;
    while (current != null) {
        Node nextNode = current.next;
        current.next = prev;
        prev = current;
        current = nextNode;
    }
    return prev; // New head of the reversed list
}
        \end{lstlisting}
        This code iteratively reverses the pointers of the linked list nodes. \\[1em]
        
        \textbf{Bad Assistant:} \\
        You can reverse it like this:
        \begin{lstlisting}
while(node != null){
    // reverse the list
}
        \end{lstlisting}
        
        \textbf{Explanation:} \\
        The Bad Assistant provides an incomplete and vague code snippet without any explanation, which is not helpful for the user trying to understand how to implement the reversal.
    \end{tcolorbox}
    \caption{Java in-context example demonstrating good and bad assistant responses. Clear code and detailed explanations are essential for user understanding.}
    \label{fig:java_in_context}
\end{figure*}

\begin{figure*}[h]
    \centering
    \begin{tcolorbox}[colback=green!5!white, colframe=green!75!black, width=\textwidth]
        \small
        \textbf{ChatHard In-Context Example} \\[0.5em]
        \textbf{Instruction:} \\
        You should provide detailed and informative answers that fully address the user's questions, avoiding overly brief or incomplete responses. \\[1em]
        
        \textbf{Example:} \\[0.5em]
        \textbf{User:} \\
        Can you explain how photosynthesis works? \\[0.5em]
        
        \textbf{Good Assistant:} \\
        Photosynthesis is the process by which green plants, algae, and some bacteria convert light energy into chemical energy. They use sunlight to synthesize nutrients from carbon dioxide and water, producing glucose and releasing oxygen as a byproduct. \\[1em]
        
        \textbf{Bad Assistant:} \\
        Plants use sunlight to make food. \\[1em]
        
        \textbf{Explanation:} \\
        The Bad Assistant's response is too brief and lacks the necessary details to fully explain the process of photosynthesis as requested.
    \end{tcolorbox}
    \caption{ChatHard in-context example showing the importance of providing detailed and comprehensive answers to fully address user questions.}
    \label{fig:chathard_in_context}
\end{figure*}

\section{Other Forms of Density Ratio as Reward}

\subsection{Delta in Prompt Conditioning Hypothesis}
Rather than leveraging difference between Strong-over-Weak models, we can potentially leverage the difference between with and without prompt conditioning for the same model to induce preference signal. For example, we can use prompt template to provide definition of preference, and contrast that with a definition-free setup. The delta will be the gains from following the pre-conditioned preference definition. 

\begin{equation}
r_{\text{prompt-template}}(x, y) = \log \pi(y \mid \text{T}(x)) - \log \pi(y \mid x)
\label{eq:prompt-conditioning}
\end{equation}

where $\template(x)$ is a function that applies a prompt template on $x$. x is input sequence and y is output sequence. $\pi$ should be an instruction tuned model, by before preference training, so that $\pi(y \mid x)$ does not have inherent understanding of preference without prompt-conditioning.

We designed experiments that set $\pi$ either as a SFT model \textit{OpenHermes-2.5-Mistral-7B} or an aligned model \textit{Nous-Hermes-2-Mistral-7B-DPO}. We then computed their reward based on \eqref{eq:prompt-conditioning}. We find that prompting only yields signal for the conditioned domain, while the other domains unrelated with conditioned prompt gives poor performance. For example, using the safety instruction in Figure~\ref{fig:safety}, $r_{\text{safety-template}}$ yields a safety score of 82.3 on RewardBench, but all other reward domains suffered, only scoring between 50-58. The overall performance is far away from safety instructed \methodname in~\eqref{eq:log_ratio-prompted} that not only boosts safety domain, but also maintain or even improve other domains' performance after. \citet{Liu2024TISDPOTI} also tries a similar setup in its \textit{TIS-DPO(P)} setup using the difference in probability between positively-prompted vs negatively-prompted sequences for importance sampling. Their negative results with this setup also confirms our negative results from simply using different prompt conditioning \eqref{eq:prompt-conditioning} as reward signal.

\begin{figure*}[t]
\vspace{-15pt}
  \centering
  \includegraphics[width=0.8\columnwidth]{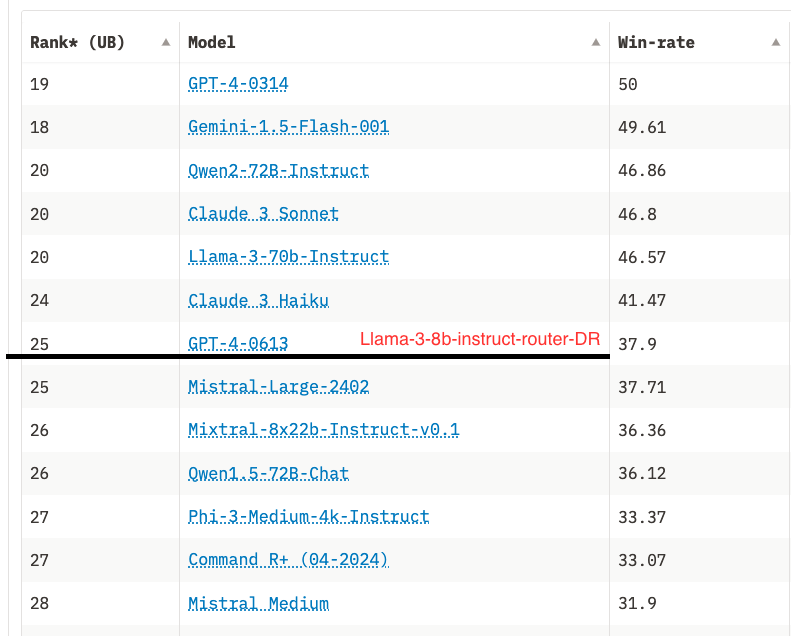}
  \caption{The ArenaHard Leaderboard. Our Llama-3-8b-instruct-router-DS stands between GPT4-0613 and Mistral-Large-2402.}
   \label{fig:arena_hard leaderboard}
\vspace{-15pt}
\end{figure*}

\end{document}